\documentclass{article} 
\usepackage{iclr2026_conference,times}

\usepackage{amsmath,amsfonts,bm}

\def\eqref#1{equation~\ref{#1}}

\def\1{\bm{1}}

\DeclareMathAlphabet{\mathsfit}{\encodingdefault}{\sfdefault}{m}{sl}
\SetMathAlphabet{\mathsfit}{bold}{\encodingdefault}{\sfdefault}{bx}{n}

\usepackage{hyperref}
\usepackage{url}
\usepackage{graphicx}
\usepackage{booktabs}
\usepackage{multirow}

\newcommand{\emailsize}{\fontsize{8.2}{10}\selectfont}

\title{Adaptive Decoding via Test-Time Policy Learning for Self-Improving Generation} 

\iclrfinalcopy

\author{Asmita Bhardwaj\thanks{This work was conducted during a Summer Internship at IBM Research.}\\
UC San Diego\\
San Diego, CA \\
\emailsize{\texttt{asbhardwaj@ucsd.edu}} \\
\And
Yuya Jeremy Ong \\
Plastic Labs \\
San Jose, CA \\
\emailsize{\texttt{yuya@plasticlabs.ai}} \\
\And
Eelaaf Zahid \\
IBM Research \\
Cambridge, MA \\
\emailsize{\texttt{eelaaf.zahid@ibm.com}} \\
\And
Basel Shbita \\
IBM Research \\
San Jose, CA \\
\emailsize{\texttt{basel@ibm.com}}}
\begin{document}

\maketitle

\begin{abstract}
Decoding strategies largely determine the quality of Large Language Model (LLM) outputs, yet widely used heuristics such as greedy or fixed temperature/top-$p$ decoding are static and often task-agnostic, leading to suboptimal or inconsistent generation quality across domains that demand stylistic or structural flexibility.
We introduce a reinforcement learning–based decoder sampler that treats decoding as sequential decision-making and learns a lightweight policy to adjust sampling parameters at test-time while keeping LLM weights frozen.
We evaluated summarization datasets including BookSum, arXiv, and WikiHow using \textit{Granite-3.3-2B} and \textit{Qwen-2.5-0.5B}.
Our policy sampler consistently outperforms greedy and static baselines, achieving relative gains of up to \textbf{+88\%} (BookSum, \textit{Granite}) and \textbf{+79\%} (WikiHow, \textit{Qwen}).
Reward ablations show that overlap-only objectives underperform compared to composite rewards, while structured shaping terms (length, coverage, repetition, completeness) enable stable and sustained improvements.
These findings highlight reinforcement learning as a practical mechanism for test-time adaptation in decoding, enabling domain-aware and user-controllable generation without retraining large models.
\end{abstract}
\section{Introduction}
Large Language Models (LLMs) have become powerful generative systems, capable of producing contextually rich text and achieving impressive performance in a wide range of tasks~\citep{naveed2025comprehensive,yang2024harnessing,wang2019superglue,brown2020language}.
Text generation involves decoding, the process of selecting output tokens from the model's predicted probability distribution.
The decoding strategy provides heuristics for token selection and plays a decisive role in shaping the final generated output.
Conventional decoding methods such as greedy sampling, top-$k$ sampling~\citep{fan2018hierarchical}, or \textit{nucleus sampling}~\citep{holtzman2019curious} rely on fixed heuristics, while other methods such as beam search~\citep{lowerre1976harpy}, contrastive decoding~\citep{Li2022ContrastiveDecoding}, or Mirostat~\citep{basu2020mirostat} introduce controlled randomness on the same principle.
Given a practically infinite search space for possible word sequences, these decoding strategies play a crucial role in the quality, diversity, and robustness of the generated text.

However, these strategies do not adapt to the evolving context of generation or the demands of different tasks, limiting the degree of freedom of the model's capability to generate text with more variations.
Often such heuristics are limited by their lack of understanding of the control capabilities and effects of the text generation process, making it unreliable and ad-hoc.
As a result, they often underperform when balancing trade-offs such as fluency, diversity, factual accuracy, and stylistic control.
Moreover, improvements in decoding behavior typically require retraining or fine-tuning the LLM itself, which is computationally expensive and impractical at scale.
From the perspective of recursive self-improvement, decoding offers a particularly attractive intervention point: it allows a model to monitor its own generation behavior, evaluate outcomes, and adjust future actions without modifying internal parameters.
Such test-time adaptation enables improvement to occur within a single episode of generation, aligning with emerging views of self-improving systems that operate through feedback, control, and policy updates rather than continual retraining.
Despite this potential, principled methods for adaptive decoding that are both lightweight and measurable remain underexplored.

We propose a learnable reinforcement learning-based decoder sampler, in which an agent learns a policy that dynamically controls and adapts decoding parameters of the LLM at test-time while keeping the LLM weights fixed.
In this context, we frame decoding as a sequential decision-making problem: the agent observes the evolving generation state (i.e., prompt, generated text, and model logits) and selects an action (i.e., setting the temperature hyperparameter) which influences the decoding behavior of the LLM.
The reward signals are derived from task-specific metrics to guide the agent toward producing higher-quality output.
This approach offers three main advantages.
First, it decouples the sampling capabilities of the decoding sampling controls from the LLM training process, which avoids shifting the underlying learned distributions of the LLM.
Second, it adapts in real time, allowing flexible control and near-real-time adaption across various tasks and domains.
Third, by formulating decoding as a Markov Decision Process (MDP)~\citep{puterman2014markov}, it provides a guiding framework for defining states, actions, and rewards that generalize beyond a single task.
Viewed through a recursive improvement lens, the proposed decoder forms a closed feedback loop: generation produces intermediate outputs, these outputs are evaluated via reward signals, and the resulting feedback informs subsequent decoding decisions within the same generation process.
While we focus on lightweight policy learning rather than long-horizon model updates, this formulation captures a core mechanism of self-improving systems under fixed model capacity.

In this work, we demonstrate the effectiveness of this method in the context of summarization, a task where decoding needs vary widely between various domains and text styles, as a demonstration of this capability.
Experiments with \textit{Granite-3.3}~\citep{granite2024granite} at 2B and \textit{Qwen-2.5}~\citep{qwen2025qwen25technicalreport} at 0.5B demonstrate consistent improvements over static baselines such as greedy decoding.
Ablation studies also reveal the contribution of different reward components and their combinations.

Our key contributions are threefold:
(i) we introduce a reinforcement learning decoder sampler that formalizes decoding as a test-time decision process with explicit state, action, and reward spaces, enabling adaptive decoding without modifying LLM weights;
(ii) we perform initial evaluations across two models and multiple datasets demonstrating gains over static strategies; and
(iii) we conduct ablation studies that highlight how different reward signals influence decoding behavior.

\section{Related Work}

Numerous decoding strategies have been developed for large language models (LLMs), encompassing both deterministic methods (e.g., beam search~\citep{lowerre1976harpy}, contrastive decoding~\citep{Li2022ContrastiveDecoding}) and stochastic approaches (e.g., top-$k$, top-$p$, Mirostat~\citep{basu2020mirostat}).
These methods primarily rely on heuristic or sampling-based mechanisms to balance fluency, diversity, and alignment with user intent.
A comprehensive taxonomy and empirical evaluation of such decoding techniques is presented by~\cite{zhang2024thorough}, highlighting both the strengths and limitations of conventional approaches in different tasks and model settings.
However, these decoding strategies are typically hand-designed and task-agnostic, making it difficult to adapt generation behavior to specific downstream objectives or contextual preferences.

To address these limitations, recent work has explored learnable and controllable decoding strategies, where decoding behavior is shaped by trainable components rather than fixed heuristics.
\cite{zhang2022survey} provide a comprehensive overview of controllable text generation using transformer-based models, highlighting methods that steer generation via prompt design, decoding constraints, or auxiliary models.
\cite{hu2025speculative} further categorize speculative and refinement-based decoding schemes that enhance generation quality or efficiency through multi-stage processes.
These works underscore a common theme: most learnable decoding approaches operate by either post-hoc control of token selection or re-ranking candidate generations. 
In contrast, our method introduces a reinforcement-learned decoder sampler that directly learns a token selection policy over a frozen LLM, enabling decoding decisions to be optimized end-to-end with respect to downstream reward signals.
This differs from prior work by tightly coupling decoding behavior with task-specific feedback, rather than relying on predefined constraints or sampling heuristics.

Our method shares conceptual similarities with ``Controlled Decoding'', which introduces a learned prefix scorer to guide decoding over a frozen base model~\cite{mudgal2024controlled}.
However, while their scorer is trained offline to approximate a static reward function, our approach learns a decoding policy directly through reinforcement learning, enabling it to receive task-specific feedback at generation time.
This allows the sampler to optimize decoding behavior end-to-end through interaction with the environment, rather than relying on precomputed supervision or reward proxies.
The reinforcement signal thus serves as a dynamic learning signal, adapting generation to complex objectives that may not be easily captured by hand-crafted scoring functions.
By combining the strengths of frozen LLMs with reinforcement-learned decoding policies, our method bridges the gap between fixed decoding heuristics and fully fine-tuned models.
\section{RL-Based Learnable Decoder Sampler}
\label{sec:decoder_sampler}

\begin{figure}[htbp]
    \centering
    \includegraphics[width=1.0\linewidth]{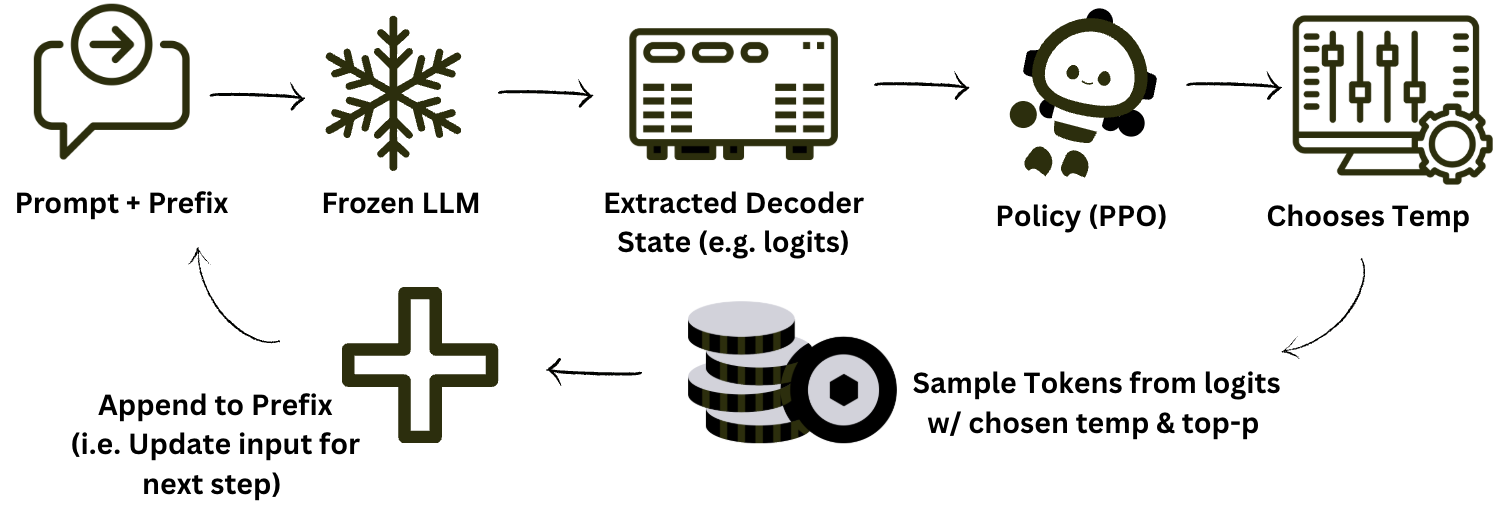}
    \caption{Overview of the RL-based decoder sampler. The agent observes the decoding state from a frozen LLM, selects decoding parameters (e.g., temperature and top-$p$), and receives rewards guiding adaptive sampling.}
    \label{fig:rl_loop}
\end{figure}

We propose to formulate decoding control as a reinforcement learning (RL) problem, where the goal is to learn an adaptive sampler that dynamically steers the output of a frozen LLM without altering its parameters.
Unlike static heuristics such as greedy search, which are hand-designed and context-agnostic, our approach introduces a learnable policy that can adapt decoding behavior to task demands in real time.
Figure~\ref{fig:rl_loop} illustrates this adaptive control loop, in which decoding decisions are continuously informed by feedback from the generated output while the underlying LLM remains fixed.

\subsection{Decoding as a Markov Decision Process}

We frame the decoding process as a Markov Decision Process (MDP).
At each decoding step, the LLM produces a probability distribution over the vocabulary, conditioned on the prompt and generated prefix.
The decoder sampler (agent) observes this evolving state and selects an action that determines how the next token is sampled.
The episode terminates when an end-of-sequence token is generated or a length limit is reached, yielding a complete feedback cycle over a single generation trajectory.

\paragraph{State.}
The state representation encapsulates information about the decoding context.
In our framework, this includes the prompt, generated prefix, model logits, and hidden representations.
Such representations provide a rich view of both the semantic history and the distributional uncertainty of the model.
In practice, we concatenate (i) mean-pooled final hidden states, (ii) top-$k$ logits (here $k=50$), (iii) prefix length, and (iv) normalized entropy of the next-token distribution.
These are linearly projected and layer-normalized to match the policy input dimension, enabling the agent to condition decisions on the unfolding generation.

\paragraph{Action.}
The action space corresponds to decoding hyper-parameters that govern token selection.
While conventional methods fix these parameters globally (e.g., setting temperature to 0.7), our agent learns to adjust them based off of context.
In this work, we focus on a two-dimensional action space controlling temperature and top-$p$.
The policy outputs Gaussian-distributed actions, which are squashed with a sigmoid and affinely mapped to parameter ranges $T\in[0.2,1.2]$ and $p\in[0.8,1.0]$.
While we restrict to these two for tractability, the framework naturally extends to additional knobs (e.g., top-$k$, length penalties, repetition penalties).

Crucially, actions taken by the policy influence not only the immediate next token but also the future state distribution through the evolving prefix, creating a closed-loop interaction between policy decisions and model behavior.
This recursive dependency allows the decoder to refine its strategy within an episode, adapting sampling behavior based on the consequences of earlier decisions.

\paragraph{Reward.} The reward function provides task-specific supervision that evaluates generated outputs and guides subsequent decoding decisions.
A detailed formulation of the reward is provided in Section~\ref{sec:experiments}.
In this work, we focus on summarization as the target task, but emphasize that summarization is not monolithic: summarizing a news article, a fictional chapter, or a Reddit post each requires different stylistic and structural trade-offs.
Static heuristics cannot easily adapt across these domains, whereas our framework is designed to flexibly incorporate reward signals that capture general qualities of a good summary (e.g., faithfulness, coverage, fluency).
This flexibility allows the same policy to adapt its decoding behavior across diverse summarization styles without hand-tuning for each case. 
However, the effectiveness of RL training critically depends on how well the reward captures the intended quality criteria; we therefore dedicate ablation studies to analyzing how different reward formulations impact policy learning and final performance (see Section~\ref{sec:experiments}).


\subsection{Policy Learning}
We employ Proximal Policy Optimization (PPO) to train the decoder sampler.
PPO offers a stable on-policy optimization procedure well-suited to low-resource RL training, where sample efficiency and stability are crucial.
The policy is parameterized as a lightweight multi-layer perceptron that maps state embeddings through a 2-layer MLP (hidden size 256, GELU activations) to a Gaussian distribution over the 2D action space.
This design ensures that the additional learnable component is compact, trainable with modest resources, and decoupled from the frozen LLM backbone.

\paragraph{Objective.}
Formally, the clipped PPO objective is:

\begin{equation}
\label{eq:ppo_obj}
\mathcal{L}_{\text{clip}}(\theta)
=\mathbb{E}_t\!\left[
\min\!\Big(
\rho_t(\theta) A_t,\;
\operatorname{clip}\!\big(\rho_t(\theta),\,1-\epsilon,\,1+\epsilon\big) A_t
\Big)
\right]
\end{equation}
where \(\rho_t(\theta) = \frac{\pi_\theta(a_t \mid s_t)}{\pi_{\theta_{\text{old}}}(a_t \mid s_t)}\) is the probability ratio and \(A_t\) is the advantage.
We estimate advantages with GAE($\gamma=0.99, \lambda=0.95$) for variance reduction. 

The full loss adds value and entropy terms:

\begin{equation}
\label{eq:loss}
\mathcal{J}(\theta)
= \mathcal{L}_{\text{clip}}(\theta)
- c_v\,\mathbb{E}_t\!\big[(V_\theta(s_t)-R_t)^2\big]
+ c_{\mathrm{ent}}\,\mathbb{E}_t\!\big[\mathcal{H}(\pi_\theta(\cdot\mid s_t))\big]
\end{equation}
and in practice we minimize \(-\mathcal{J}(\theta)\).


\subsection{Framework Flexibility}
A key advantage of our approach is modularity: the same RL formulation can be instantiated with different state features, action knobs, and reward functions depending on the target application.
For example, abstractive summarization may benefit from rewards emphasizing conciseness and factual coverage, whereas story generation could prioritize diversity and creativity.
By learning a decoding policy end-to-end, the sampler adapts dynamically to input prompts and domain variation, moving beyond static, one-size-fits-all heuristics.
This modularity makes the decoder sampler a natural substrate for recursive self-improvement at test-time, where evaluation signals can be iteratively incorporated without modifying model parameters.

In Section~\ref{sec:experiments}, we instantiate this framework on summarization tasks using \textit{Granite-3.3-2B-Base} and \textit{Qwen-2.5-0.5B}, comparing against conventional baselines and analyzing the impact of different reward designs. 
\section{Evaluation}
\label{sec:experiments}

\subsection{Experimental Setup}

We evaluate the proposed RL-based decoder sampler on abstractive summarization, a natural testbed that spans diverse domains and styles. This setting allows us to assess whether adaptive sampling improves robustness across tasks where static decoding heuristics are commonly brittle.
This evaluation targets a constrained but measurable form of self-improvement: a lightweight controller that updates its behavior through feedback while the base model remains unchanged.
We therefore emphasize (i) within-episode adaptation (state-dependent decoding actions) and (ii) across-training improvement (early$\to$late PPO gains) as complementary evidence that the system is not merely a re-parameterized heuristic.

\paragraph{Datasets.}
We use three domains:
(1) \textbf{BookSum} (narrative chapters, long-form);
(2) \textbf{arXiv} (scientific articles/abstracts); and
(3) \textbf{WikiHow} (instructional ``how-to''). 
These datasets differ substantially in length, factuality requirements, and stylistic variation, providing a broad stress test for adaptive decoding.

\paragraph{Models.}
We evaluate \textbf{\textit{Granite-3.3}~(2B Base)} and \textbf{\textit{Qwen-2.5}~(0.5B)}. The LLMs remain frozen throughout; only the lightweight policy network is trained.

\paragraph{Training details.}
Each PPO run uses \textbf{100 sampled prompts per dataset}, reflecting a low-resource RL setting.
For each prompt we generate a candidate summary with the current policy, compute the reward, and update periodically.
Unless noted otherwise, prompts are sampled without overlap between training and evaluation splits to avoid overestimating reward-driven adaptation.
We report both (i) absolute average rewards and (ii) \emph{early$\to$late} changes to show whether PPO training improves policy quality rather than collapsing to a static strategy.
We define \emph{early$\to$late} as the relative change in average reward between the first and last evaluation checkpoints within a run (computed under the same reward variant), which serves as a direct indicator of whether policy learning yields improvement over training time.
Hyperparameters follow the defaults in Section~\ref{sec:decoder_sampler}: PPO with clipping parameter $\epsilon=0.2$, $\gamma=0.99$, $\lambda=0.95$, value coefficient $c_v=0.5$, and entropy coefficient $c_{\mathrm{ent}}=0.01$.

\paragraph{Baselines.}
We compare against two common heuristics:
\begin{itemize}
    \item \emph{Greedy decoding}: argmax at each step.
    \item \emph{Static sampling}: fixed temperature of 0.3, held constant across prompts.
\end{itemize}
We choose a conservative static temperature to represent a common ``safe'' setting for summarization that typically trades diversity for precision; the RL policy is evaluated under the same decoding infrastructure for a controlled comparison.
These baselines provide strong static references for assessing whether learning a decoding policy yields tangible benefits.

\paragraph{Rewards.}
Our \textbf{Proposed Reward} is defined as a weighted, normalized combination of:
\begin{itemize}
    \item \emph{ROUGE-L F1} (weight \(0.5\)): overlap with reference summaries.
    \item \emph{Length penalty} (weight \(0.2\)): deviation from an ideal source-derived length.
    \item \emph{Coverage bonus} (weight \(0.1\), capped): overlap with important source tokens (\(>4\) chars).
    \item \emph{Repetition penalty}: applied if more than \(30\%\) of words are repeated.
    \item \emph{Completeness penalty} (\(-0.05\)): applied if the summary lacks sentence-ending punctuation.
\end{itemize}

The raw score is linearly normalized to \([0,1]\) via \(\texttt{RAW\_MIN}, \texttt{RAW\_MAX}\).  

This composite reward is designed to balance multiple, sometimes competing, aspects of summary quality, rather than optimizing a single metric in isolation.
Because reinforcement learning is highly sensitive to the shape and scale of the reward signal, small design choices can substantially influence both training stability and final behavior.
To better understand these effects, we systematically ablate individual reward components as described next.

We perform several ablations to isolate the impact of reward design:
\begin{itemize}
    \item \textbf{ROUGE-only}: remove all shaping terms (length, coverage, repetition, completeness).
    \item \textbf{Core shaping}: keep ROUGE + length + repetition, drop coverage.
    \item \textbf{Proposed w/o Coverage}: remove the coverage bonus term only.
    \item \textbf{Proposed w/ Softer Repetition}: reduce the weight of the repetition penalty.
    \item \textbf{Proposed w/ Sigmoid Scaling}: replace the linear normalization with a sigmoid squashing.
\end{itemize}

Because reward scales differ, we compare results in relative terms:  
(i) PPO vs.\ baselines under the \emph{same} reward, and  
(ii) PPO \emph{early$\to$late} change.
This protocol isolates the role of evaluation signals in enabling test-time controllability: changing the reward changes what the controller learns to optimize, without altering the base LLM.


\subsection{Results and Discussion}

\begin{figure}[htbp]
\centering
\includegraphics[width=0.56\linewidth]{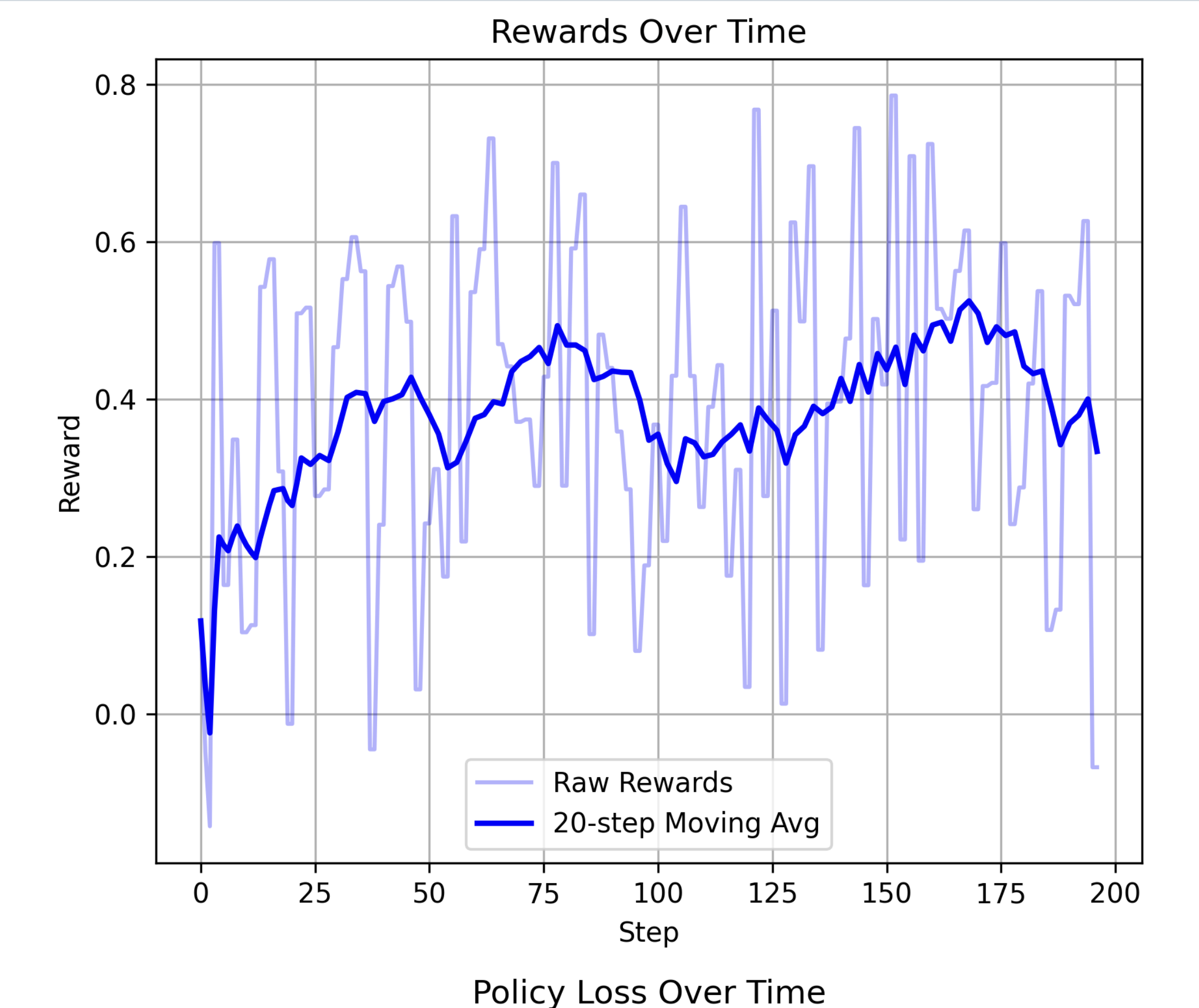}
\caption{\textbf{Reward over time} for WikiHow + \textit{Granite-3.3} under the Proposed Reward. The upward trend in the moving average demonstrates PPO learns a useful sampling policy.}
\label{fig:train_wikihow_granite}
\end{figure}

\paragraph{Training signal (proof of learning).}
Before comparing against baselines, we first verify that PPO training yields meaningful improvements. 
Figure~\ref{fig:train_wikihow_granite} shows the reward trajectory for \textbf{WikiHow + \textit{Granite-3.3}} under the Proposed Reward. 
The moving average rises steadily during training, confirming that the policy learns to improve generation quality rather than collapsing to a static heuristic.
This provides empirical evidence that the feedback loop in Figure~\ref{fig:rl_loop} supports learnable, non-trivial control over decoding behavior.

\paragraph{Main results (comparison to baselines).}
Table~\ref{tab:main_results} compares PPO-trained policies against greedy and static baselines across all datasets and models. 
RL consistently outperforms heuristics, with especially large improvements on BookSum and WikiHow, where stylistic diversity and length variation make fixed parameters brittle. 
Early$\to$late changes are generally modest but positive, confirming that PPO training improves policy quality rather than collapsing to static strategies.

\begin{table*}[ht]
   \centering
   \caption{Main evaluation results with the Proposed Reward. RL achieves consistent improvements over greedy and static baselines. Scores are reported alongside percentage gains and PPO early$\to$late improvements.}
   \label{tab:main_results}
   \begin{tabular}{lrrrrrr}
   \toprule
    & \multicolumn{3}{c}{Granite-3.3-2B} & \multicolumn{3}{c}{Qwen-2.5-0.5B} \\ 
    \cmidrule(lr){2-4} \cmidrule(lr){5-7}
    & \rotatebox{90}{WikiHow} & \rotatebox{90}{arXiv} & \rotatebox{90}{BookSum}  & 
    \rotatebox{90}{WikiHow} & \rotatebox{90}{arXiv} & \rotatebox{90}{BookSum} \\
   \midrule
   Greedy                      & 0.333  & 0.220  & 0.147  & 0.114  & 0.154  & 0.014   \\
   Static                      & 0.369  & 0.275  & 0.212  & 0.140  & 0.174  & 0.083   \\
   RL (Proposed)               & \textbf{0.390}  & \textbf{0.353}  & \textbf{0.277}  & \textbf{0.204}  & \textbf{0.226}  & \textbf{0.147}   \\
   \midrule
   $\Delta$ vs Greedy (\%)     & +17.17 & +60.45 & +88.44 & +78.95 & +46.75 & +950.00  \\
   $\Delta$ vs Static (\%)     & +5.69  & +28.36 & +30.66 & +45.71 & +29.89 & +77.11  \\
   Early$\to$Late (\%)         & +22.88 & -4.38  & +22.06 & -43.60 & +54.09 & +57.14  \\
   \bottomrule
   \end{tabular}
\end{table*}

\paragraph{Reward ablations (impact of design).}
Table~\ref{tab:ablations} focuses on a representative case, \textbf{arXiv + \textit{Granite-3.3}}, to isolate the effect of reward design. 
ROUGE-only yields negligible or even negative gains, confirming that overlap metrics alone are insufficient. 
Core shaping improves over baselines but shows instability. 
Removing coverage produces large gains with positive early$\to$late improvements, while sigmoid scaling is modest but stable. 
Our Proposed Reward balances strong gains with robustness, supporting our claim that structured shaping terms are critical.

\begin{table}[htbp]
\centering
\caption{Reward ablations on \textbf{arXiv (\textit{Granite-3.3-2B})}. We report absolute rewards, percentage improvements over baselines, and PPO early$\to$late change.}
\label{tab:ablations}
\begin{tabular}{lrrrrrr}
\toprule
Reward Variant & \rotatebox{90}{PPO} & \rotatebox{90}{Greedy} & \rotatebox{90}{Static} & \rotatebox{90}{$\Delta$ vs Greedy (\%)} & \rotatebox{90}{$\Delta$ vs Static (\%)} & \rotatebox{90}{Early$\to$Late (\%)} \\
\midrule
ROUGE-only                & 0.169 & 0.158 & 0.170 & +6.96  & -0.59  & +15.38 \\
Core shaping              & 0.448 & 0.335 & 0.400 & +33.73 & +12.00 & -4.56  \\
No coverage bonus         & 0.349 & 0.219 & 0.275 & +59.36 & +26.91 & +24.91 \\
Soft repetition penalty   & 0.503 & 0.431 & 0.462 & +16.72 &  +8.88 & +6.81  \\
Sigmoid scaling           & 0.382 & 0.302 & 0.323 & +26.49 & +18.27 & -4.27  \\
\midrule
Proposed (ours)           & 0.353 & 0.220 & 0.275 & +60.45 & +28.36 & -4.38  \\
\bottomrule
\end{tabular}
\end{table}

\paragraph{Takeaways.}
Our findings yield several insights into the effectiveness of RL-based decoding and the role of reward design:

\begin{enumerate}
    \item \textbf{PPO training is stable and effective.}  
    The reward-over-time curve (Figure~\ref{fig:train_wikihow_granite}) demonstrates that PPO steadily increases reward under the Proposed Reward, confirming the agent learns a non-trivial policy rather than collapsing to a static strategy.

    \item \textbf{RL policies outperform static heuristics.}  
    Across datasets and models (Table~\ref{tab:main_results}), PPO-trained policies consistently achieve higher rewards than greedy and static decoding. The magnitude of these gains depends on the domain: in stylistically diverse tasks like BookSum and WikiHow, PPO delivers substantial improvements (e.g., $+88\%$ on BookSum \textit{Granite-3.3}; $+79\%$ on WikiHow \textit{Qwen-2.5}), showing that adaptive parameter control is particularly valuable where one-size-fits-all heuristics fail. Even where improvements are modest (e.g., arXiv), PPO at least matches static strategies without degradation, suggesting robustness across domains.

    \item \textbf{Reward design is critical.}  
    The ablation study (Table~\ref{tab:ablations}) reveals that reward shaping strongly influences learning outcomes. ROUGE-only rewards yield minimal gains and sometimes underperform static baselines, confirming that overlap alone is insufficient to capture summary quality. Adding shaping terms improves performance but affects stability differently:  
    (i) Core shaping boosts absolute reward but exhibits negative early$\to$late trends, indicating instability;
    (ii) Removing coverage produces large, stable gains, suggesting coverage is not always beneficial and may introduce noise;
    (iii) Softer repetition penalties dampen improvements, showing that repetition control must be sufficiently strict; and
    (iv) Sigmoid scaling smooths rewards but reduces training sensitivity, leading to moderate but stable results;
    Our Proposed Reward achieves a balance of gains and robustness, supporting the claim that well-designed, multi-component rewards are necessary for controllable generation.

    \item \textbf{Domain effects.}  
    Improvements vary with domain. BookSum and WikiHow, which demand stylistic flexibility and length adaptation, benefit most from adaptive decoding. In contrast, arXiv summaries are more formulaic and yield smaller but consistent gains, showing that RL-based control is most useful when the target style diverges from the assumptions of fixed heuristics. This suggests that adaptive samplers could be especially impactful in open-domain or user-facing applications where prompt style varies widely.

    \item \textbf{Model effects.}  
    Even small models like \textit{Qwen-2.5-0.5B} benefit significantly from adaptive sampling. Relative gains are often larger than for \textit{Granite-3.3-2B}, demonstrating that RL-based decoding control is not limited to large-scale LLMs and may be even more impactful in constrained settings. This broadens the applicability of our approach to scenarios where fine-tuning is infeasible but decoding-time adaptation can deliver tangible improvements.

    \item \textbf{Latency and deployment overhead.} The policy is a 2-layer MLP with hidden size 256, so per-token compute is negligible relative to a full LLM forward pass. The primary state features are already produced during standard decoding and do not require additional forward passes. For resource-constrained settings, either k in the top-k logit features or hidden-state inputs can be reduced at a potential cost to policy expressiveness.

    \item \textbf{Broader implications.} RL-based samplers can reliably improve generation quality without retraining the LLM, and reward design is as important as the RL algorithm itself. Rather than hand-crafting decoding heuristics, lightweight agents can learn domain-specific strategies guided by modular reward definitions.
    
\end{enumerate}

We note that this study targets short-horizon improvement in a bounded setting (summarization) rather than long-horizon autonomous self-modification, but it provides a concrete and measurable stepping stone toward loop-based systems. 
\section{Conclusion}
We introduced a reinforcement learning–based decoder sampler that adaptively adjusts decoding parameters of frozen LLMs at inference time.
By framing decoding as a sequential decision-making problem, our method learns lightweight policies that improve generation quality without requiring model fine-tuning. 
Experiments across two models and three datasets show PPO-trained samplers consistently outperforming greedy and static baselines, with reward ablations confirming that structured shaping terms are critical for stable learning.

These findings underscore two key themes: reinforcement learning is a practical tool for controllable decoding, and reward design is as important as the RL algorithm itself.
Looking ahead, we envision extending this framework to richer action spaces (e.g., top-$k$, top-$p$, penalties), exploring preference-based or LLM-as-a-judge reward signals, and applying adaptive decoding to broader generative tasks such as dialogue or story generation.
By decoupling decoding control from model parameters, we hope this line of work opens new pathways toward flexible, domain-aware, and user-controllable language generation.

All experiments target summarization; transfer to dialogue, code generation, or reasoning, where reward specification is harder, remains untested. The evaluated models (0.5B–2B) represent the lower end of current LLM scale, and out-of-distribution robustness of the learned policy is an open empirical question. We identify both as directions for future work.
 
Since the controller operates at inference time and does not modify model parameters, it avoids permanent regressions to the base model and enables straightforward rollback by disabling the learned policy.
Moreover, policy behavior can be monitored and constrained through explicit reward definitions and action bounds, providing a clear surface for auditing and governance.
Together, these properties suggest that test-time decoder adaptation offers a lower-risk pathway toward self-improving generation compared to weight-level updates, particularly in settings where stability and reversibility are critical.


\bibliography{iclr2026_conference}
\bibliographystyle{iclr2026_conference}


\end{document}